%% file: main.tex
\documentclass[11pt,a4paper]{article}
\usepackage{authblk}
\usepackage[hyperref]{acl2019}
\usepackage{times}
\usepackage{latexsym}
\usepackage{graphicx}
\usepackage{amsmath}
\usepackage{booktabs}
\usepackage{algorithm}
\usepackage{algorithmic}

\usepackage[flushleft]{threeparttable}
\usepackage{verbatim}
\usepackage{multicol}
\usepackage{multirow}
\usepackage{subfigure}
\usepackage{float}
\usepackage{mathrsfs}
\usepackage{color}
\usepackage{todonotes}
\usepackage{soul}

\usepackage{url}
\usepackage{xspace}
\usepackage{ amssymb }

\newcommand{\fullmethod}{Metropolis-Hastings attack\xspace}
\newcommand{\method}{MHA\xspace}
\newcommand{\bbmethod}{\textit{b}-MHA\xspace}
\newcommand{\wbmethod}{\textit{w}-MHA\xspace}

\aclfinalcopy 
\def\aclpaperid{2666} 

\title{Generating Fluent Adversarial Examples for Natural Languages}

\author{\textbf{Huangzhao Zhang}$^{1}$\thanks{~~~Work done while Huangzhao Zhang was a research intern in
ByteDance AI Lab, Beijing, China.} \quad
\textbf{Hao Zhou}$^{2}$ \quad
\textbf{Ning Miao}$^{2}$ \quad
\textbf{Lei Li}$^{2}$ \\ 
$^{1}$Institute of Computer Science and Technology, Peking University, China\\
$^{2}$ByteDance AI Lab, Beijing, China \\
\texttt{zhang\_hz@pku.edu.cn}\ \\
\texttt{\{miaoning,zhouhao.nlp,lileilab\}@bytedance.com}
}

\date{}

\begin{document}
\maketitle

\begin{abstract}
Efficiently building an adversarial attacker for natural language processing~(NLP) tasks is a real challenge. Firstly, as the sentence space is discrete, it is difficult to make small perturbations along the direction of gradients. Secondly, the fluency of the generated examples cannot be guaranteed. In this paper, we propose \method, which addresses both problems by performing Metropolis-Hastings sampling, whose proposal is designed with the guidance of gradients. Experiments on IMDB and SNLI show that
our proposed \method outperforms the baseline model on attacking capability. Adversarial training with \method also leads to better robustness and performance.
\end{abstract}

\input{introduction}
\input{definition}
\input{methods}
\input{experiments}
\input{summary}


\bibliographystyle{acl_natbib}
\bibliography{reference}

\end{document}


\maketitle

\appendix

\begin{table*}[b!]
    \centering
    \begin{threeparttable}
    \begin{tabular}{p{\textwidth}}
        \toprule
\multicolumn{1}{c}{Examples} \\
        \midrule
        
\textbf{Premise:} \emph{a group of people examine a boat with an orange flag that is sitting on sand next to a body of water.} \\
\textbf{Original hypothesis:} \emph{people sit on a beach to tan.} \textbf{Prediction:} $\langle$Neutral$\rangle$ \\
\textbf{Genetic hypothesis:} \emph{people sit on a \textbf{\color{red}swimming} to tan.} \textbf{Prediction:} $\langle$Contradiction$\rangle$ \\
\textbf{\bbmethod hypothesis:} \emph{people sit on a beach \textbf{\color{brown}and} tan.} \textbf{Prediction:} $\langle$Contradiction$\rangle$ \\
\textbf{\wbmethod hypothesis:} \emph{people sit on a beach \textbf{\color{blue}and} tan.} \textbf{Prediction:} $\langle$Contradiction$\rangle$ \\
        
\cline{1-1}
\textbf{Premise:} \emph{a woman lying in the grass in the park is wearing a red top and black capri pants and is barefoot.} \\
\textbf{Original hypothesis:} \emph{a woman is sitting on a park bench wearing sandals.} \textbf{Prediction:} $\langle$Contradiction$\rangle$ \\
\textbf{Genetic hypothesis:} \emph{a woman is \textbf{\color{red}seated} on a park bench wearing \textbf{\color{red}footwear}.} \textbf{Prediction:} $\langle$Entailment$\rangle$ \\
\textbf{\bbmethod hypothesis:} \emph{a woman is sitting on a park bench wearing \textbf{\color{brown}it}.} \textbf{Prediction:} $\langle$Entailment$\rangle$ \\
\textbf{\wbmethod hypothesis:} \emph{a woman is sitting on a park bench wearing \textbf{\color{blue}it}.} \textbf{Prediction:} $\langle$Entailment$\rangle$ \\

\cline{1-1}
\textbf{Premise:} \emph{a man alone crosscountry skis in the wilderness while wearing a huge backpack.} \\
\textbf{Original hypothesis:} \emph{a man skis in the wilderness while it's snowing} \textbf{Prediction:} $\langle$Neutral$\rangle$ \\
\textbf{Genetic hypothesis:} \emph{a man \textbf{\color{red}snowboarding} in the wilderness while it's snowing} \textbf{Prediction:} $\langle$Contradiction$\rangle$ \\
\textbf{\bbmethod hypothesis:} \emph{a man skis in the \textbf{\color{brown}world} while it's snowing} \textbf{Prediction:} $\langle$Contradiction$\rangle$ \\
\textbf{\wbmethod hypothesis:} \emph{a man skis in the wilderness \textbf{\color{blue}and} it's snowing} \textbf{Prediction:} $\langle$Contradiction$\rangle$ \\

\cline{1-1}
\textbf{Premise:} \emph{a boy kneeling on a skateboard riding down the street} \\
\textbf{Original hypothesis:} \emph{a boy standing upright on a skateboard.} \textbf{Prediction:} $\langle$Contradiction$\rangle$ \\
\textbf{Genetic hypothesis:} \emph{a boy \textbf{\color{red}permanent} upright on a skateboard.} \textbf{Prediction:} $\langle$Entailment$\rangle$ \\
\textbf{\bbmethod hypothesis:} \emph{a boy \textbf{\color{brown}is} \textbf{\color{brown}out} on a skateboard .} \textbf{Prediction:} $\langle$Entailment$\rangle$ \\
\textbf{\wbmethod hypothesis:} \emph{a boy \textbf{\color{blue}was} upright on a skateboard.} \textbf{Prediction:} $\langle$Entailment$\rangle$ \\

\cline{1-1}
\textbf{Premise:} \emph{three men are sitting on a beach dressed in orange with refuse carts in front of them.} \\
\textbf{Original hypothesis:} \emph{empty trash cans are sitting on a beach.} \textbf{Prediction:} $\langle$Contradiction$\rangle$ \\
\textbf{Genetic hypothesis:} \emph{\textbf{\color{red}empties} trash cans are sitting on a beach..} \textbf{Prediction:} $\langle$Entailment$\rangle$ \\
\textbf{\bbmethod hypothesis:} \emph{\textbf{\color{brown}the} trash cans are sitting \textbf{\color{brown}in} a beach.} \textbf{Prediction:} $\langle$Entailment$\rangle$ \\
\textbf{\wbmethod hypothesis:} \emph{\textbf{\color{blue}the} trash cans are sitting on a beach.} \textbf{Prediction:} $\langle$Entailment$\rangle$ \\

\cline{1-1}
\textbf{Premise:} \emph{hikers walk along some tough terrain.} \\
\textbf{Original hypothesis:} \emph{hiking pace along rough terrain.} \textbf{Prediction:} $\langle$Entailment$\rangle$ \\
\textbf{Genetic hypothesis:} \emph{hiking pace along rough \textbf{\color{red}terra}.} \textbf{Prediction:} $\langle$Neutral$\rangle$ \\
\textbf{\bbmethod hypothesis:} \emph{hiking \textbf{\color{brown}is} \textbf{\color{brown}in} rough terrain.} \textbf{Prediction:} $\langle$Neutral$\rangle$ \\
\textbf{\wbmethod hypothesis:} \emph{\textbf{\color{blue}the} pace along rough terrain.} \textbf{Prediction:} $\langle$Neutral$\rangle$ \\

\cline{1-1}
\textbf{Premise:} \emph{our people walking beside each other down a street, one of the men is turned around looking toward the camera.} \\
\textbf{Original hypothesis:} \emph{a group of friends are headed to wendys} \textbf{Prediction:} $\langle$Neutral$\rangle$ \\
\textbf{Genetic hypothesis:} \emph{a groups of \textbf{\color{red}boyfriends} are \textbf{\color{red}guided} to wendys} \textbf{Prediction:} $\langle$Contradiction$\rangle$ \\
\textbf{\bbmethod hypothesis:} \emph{a group of \textbf{\color{brown}women} are \textbf{\color{brown}expected} to wendys} \textbf{Prediction:} $\langle$Contradiction$\rangle$ \\
\textbf{\wbmethod hypothesis:} \emph{a \textbf{\color{blue}number} of \textbf{\color{blue}people} are \textbf{\color{blue}going} to wendys} \textbf{Prediction:} $\langle$Contradiction$\rangle$ \\

\cline{1-1}
\textbf{Premise:} \emph{a man in a green shirt hovers above the ground in the laundry room.} \\
\textbf{Original hypothesis:} \emph{the man appears to be suspended in midair.} \textbf{Prediction:} $\langle$Entailment$\rangle$ \\
\textbf{Genetic hypothesis:} \emph{the man \textbf{\color{red}emerge} to be suspended in midair.} \textbf{Prediction:} $\langle$Neutral$\rangle$ \\
\textbf{\bbmethod hypothesis:} \emph{the man appears to be suspended in \textbf{\color{brown}2007}.} \textbf{Prediction:} $\langle$Neutral$\rangle$ \\
\textbf{\wbmethod hypothesis:} \emph{the man \textbf{\color{blue}is} to be suspended in midair.} \textbf{Prediction:} $\langle$Neutral$\rangle$ \\

        \bottomrule
    \end{tabular}
    \end{threeparttable}
    \caption{Adversarial examples generated on SNLI.}
    \label{table:cases}
\end{table*}

\section{Hyper-parameters}

The hyper-parameters of the \method model, the genetic baseline model, and the victim models are listed as follow.

\paragraph{\method.} We set the hyper-parameters of \method to $p_r=0.5$, $p_i=0.25$, $p_d=0.25$. Constraints on $LM(x)$ and $C(\tilde{y}|x)$ is performed -- if $LM(x')<t_{LM}\cdot LM(x)$ or $C(\tilde{y}|x')<t_C\cdot C(\tilde{y}|x)$, the proposal is rejected directly. Such trick ensures that we do not loss sentence fluency or target probability rapidly. $t_{LM}$ and $t_C$ are set to 0.8 and 0.9 in our experiments. Also, any operation on sentimental words (eg. ``great'') or negation words (eg. ``not'') are forbidden in IMDB experiments. SentiWordNet \citep{esuli2006sentiwordnet,baccianella2010sentiwordnet} are applied to recognize the sentimental words.

The language models in \method includes a forward and a backward 2-layer LSTM models with 300 units trained on subset of the One-Billion-Word Corpus \citep{chelba2013one}.
We randomly select 5M sentences from the corpus for LM training. The vocabulary size is 50,000. The two LSTMs employ independent embedding matrices with the same word2vec initialization.

\paragraph{Genetic baseline.} The hyper-parameter settings of the genetic model remain the same as in the paper \citep{alzantot2018generating}.

\paragraph{Victim models.} The LSTMs in the victim models have 128 units. The bi-LSTM for IMDB has a vocabulary size of 10,000, while the two LSTMs in the BiDAF model share the same vocabulary size of 35,000. The embedding matrices are pre-trained by word2vec. In addition, the embedding matrix of the bi-LSTM model for IMDB is fixed during training to avoid overfitting. All classifiers in our experiment reach 99\% accuracy on the training set.

\section{Adversarial Examples}

Some adversarial examples are listed in Table \ref{table:cases}. The genetic replacement considers only the current word itself, regardless of its context, and results in an unfluent sentence, while \method performs replacement with the guide of LM, and the sentence is fluent.

Empirically, \method is allowed to operates all types of words, including the prepositions, the pronouns, and the punctuations, \emph{etc.}, where these changes are minor. While the genetic approach only replace the verbs, the nouns, the adjectives and the adverbs. 
An advantage of operating the prepositions, the punctuations, \emph{etc.} is that human beings usually do not pay much attention to them. Human begins can hardly recognize the adversarial examples generated by these operations.

\bibliographystyle{acl_natbib}
\bibliography{reference}

%% file: introduction.tex
\section{Introduction}
\label{section:intro}

Adversarial learning has been a popular topic in deep learning. Attackers generate adversarial examples by perturbing the samples and use these examples to fool deep neural networks~(DNNs).
From the perspective of defense, adversarial examples are mixed into the training set to improve performance and robustness of the victim models.

\begin{figure}[t]
    \centering
    \subfigure[Adversarial training with fluent adversarial examples]{
        \includegraphics[width=0.45\columnwidth]{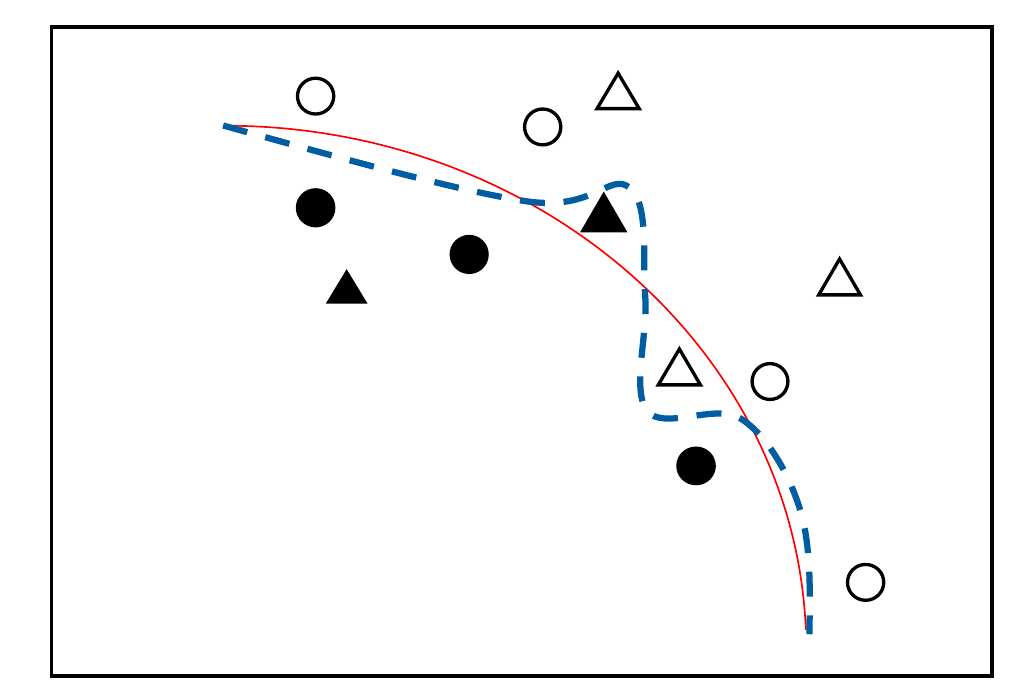}
    }
    \subfigure[Adversarial training with unfluent adversarial examples]{
        \includegraphics[width=0.45\columnwidth]{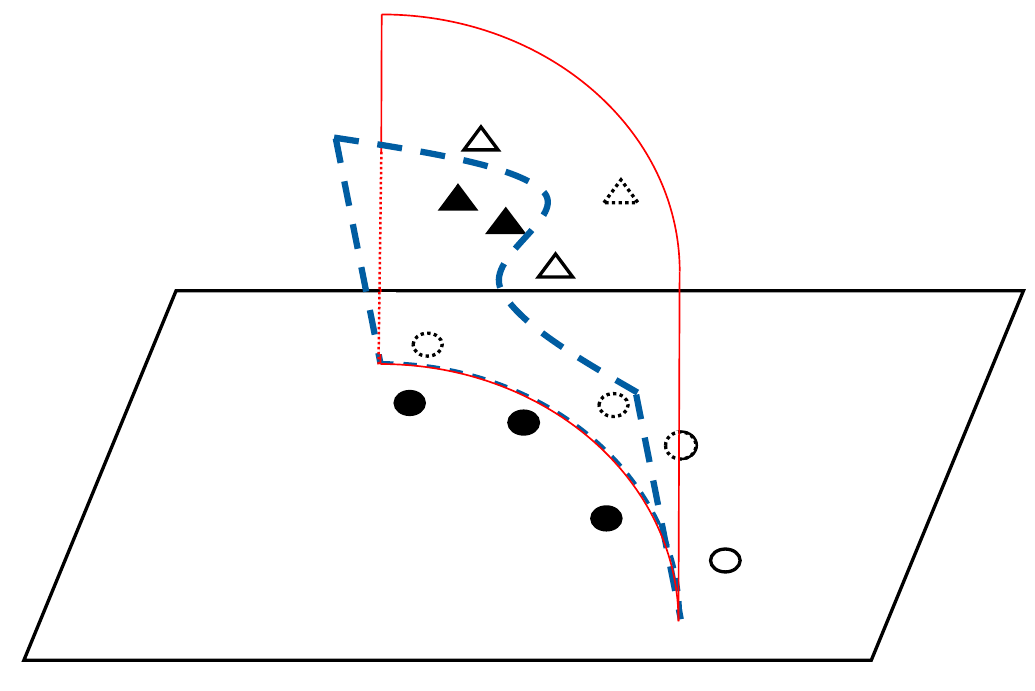}
    }
    \caption{Effect of adversarial training on (a) fluent and (b) unfluent adversarial examples. $\circ$ and $\bullet$ represent positive and negative samples in the training set, while $\vartriangle$ and $\blacktriangle$ are the corresponding adversarial examples. Solid and Dotted lines represent decision boundaries before and after adversarial training, respectively. As unfluent adversarial examples are not in the manifold of real sentences, the victim model only needs to adjust its decision boundary out of the sentence manifold to fit them. As a result, fluent adversarial examples may be more effective than unfluent ones.}
    \label{figure:manifold}
\end{figure}

However, building an attacker for NLP models (such as a text classifier) is extremely challenging. Firstly, it is difficult to perform gradient-based perturbations since the sentence space is discrete. However, gradient information is critical -- it leads to the steepest direction to more effective examples.
Secondly, adversarial examples are usually not fluent sentences. Unfluent examples are less effective in attacking, as victim models can easily learn to recognize them. Meanwhile, adversarial training on them usually does not perform well (see Figure \ref{figure:manifold} for detailed analysis). 

Current methods cannot properly handle the two problems. \citet{ebrahimi2018hotflip} (HotFlip) propose to perturb a sentence by flipping one of the characters, and use the gradient of each perturbation to guide sample selection. But simple character flipping often leads to meaningless words (\emph{eg.} ``moo\textbf{d}'' to ``moo\textbf{P}''). Genetic attack \citep{alzantot2018generating} is a population-based word replacing attacker, which aims to generate fluent sentences by filtering out the unreasonable sentences with a language model. But the fluency of examples generated by genetic attack is still not satisfactory and it is inefficient as the gradient is discarded.

To address the aforementioned problems, we propose the \fullmethod (\method) algorithm in this short paper. \method is an adversarial example generator based on Metropolis-Hastings (M-H) sampling \citep{metropolis1953equation,hastings1970monte,chib1995understanding}. M-H sampling is a classical MCMC sampling approach, which has been applied to many NLP tasks, such as natural language generation \citep{kumagai2016human}, constrained sentence generation~\citep{miao2018cgmh}, guided open story generation~\citep{harrison2017toward}, \emph{etc.}
We propose two variants of MHA, namely a black-box \method (\bbmethod) and a white-box \method (\wbmethod).
Specifically, in contrast to previous language generation models using M-H, \bbmethod's stationary distribution is equipped with a language model term and an adversarial attacking term.
The two terms make the generation of adversarial examples fluent and effective.  
\wbmethod even incorporates adversarial gradients into proposal distributions to speed up the generation of adversarial examples.

Our contributions include that we propose an efficient approach for generating fluent adversarial examples.
Experimental results on IMDB~\citep{maas2011learning} and SNLI~
\citep{bowman2015large} show that, compared with the state-of-the-art genetic model, \method generates examples faster, achieving higher success rates with much fewer invocations. 
Meanwhile, adversarial samples from \method are not only more fluent but also more effective to improve the adversarial robustness and classification accuracy after adversarial training.


%% file: definition.tex
\section{Preliminary}
\label{section:definition}

\begin{figure}[t]
    \centering
    \includegraphics[width=0.9\columnwidth]{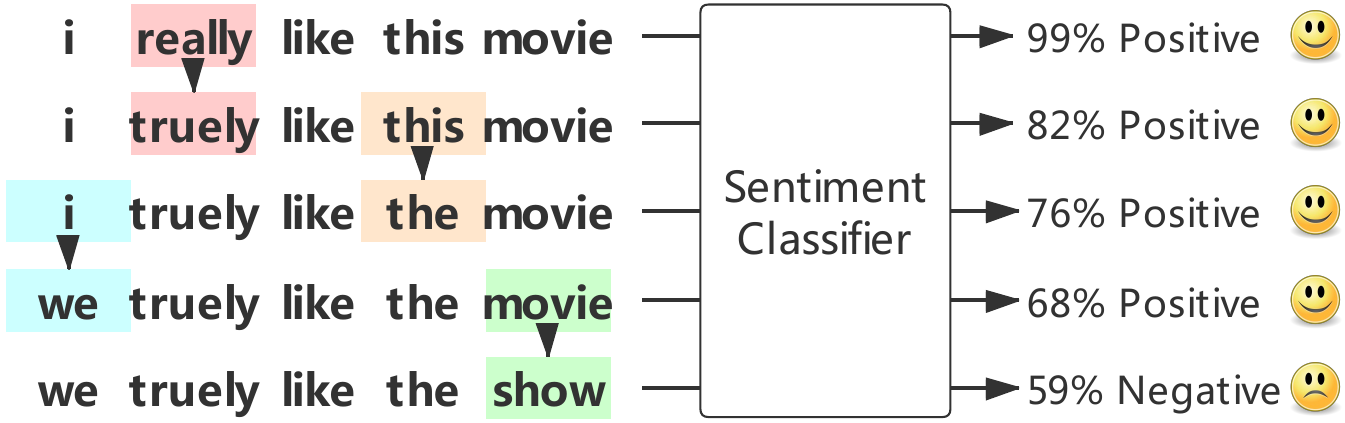}
    \caption{A simple example of adversarial attack on a sentimental classifier by performing word replacement.}
    \label{figure:attack}
\end{figure}

Generally, adversarial attacks aim to mislead the neural models by feeding adversarial examples with perturbations, while adversarial training aims to improve the models by utilizing the perturbed examples.
Adversarial examples fool the model into producing erroneous outputs, such as irrelevant answers in QA systems or wrong labels in text classifiers~(Figure \ref{figure:attack}).
Training with such examples may enhance performance and robustness.

Definitions of the terms in this paper are as follow. The \textbf{victim models} are word-level classifiers, which take in tokenized sentences and output their labels. 
The \textbf{attackers} generate sentences by perturbing the original ones, in order to mislead the victim model into making mistakes. 
\textbf{Adversarial attacks} include two categories: (a) \textbf{black-box attack} only allows the attackers to have access to model outputs,
while (b) \textbf{white-box attack} allows full access to the victim model, including model outputs, gradients and (hyper-)parameters. 
For \textbf{adversarial training}, the same victim model is trained from scratch on an updated training set with adversarial examples included.


%% file: methods.tex
\section{Proposed Method: \method}
\label{section:methods}

In this section, we first introduce M-H sampling briefly, and then describe how to apply M-H sampling efficiently to generate adversarial examples for natural language.

\subsection{Metropolis-Hastings Sampling}
\label{section:methods:subsection:mh}


The M-H algorithm is a classical Markov chain Monte Carlo sampling approach. Given the stationary distribution ($\pi(x)$) and transition proposal, M-H is able to generate desirable examples from $\pi(x)$. Specifically, at each iteration, a proposal to jump from $x$ to $x'$ is made based on the proposal distribution ($g(x'|x)$). The proposal is accepted with a probability given by the acceptance rate:
\begin{equation}
    \alpha(x'|x)=\min\{1,\frac{\pi(x')g(x|x')}{\pi(x)g(x'|x)}\} 
\end{equation}
Once accepted, the algorithm jumps to $x'$. Otherwise, it stays at $x$.

\subsection{Black-Box Attack}
\label{section:methods:subsection:mh-bb}

In black-box attack (\bbmethod), we expect the examples to meet three requirements: (a) to read fluently; (b) to be able to fool the classifier; (c) to invoke the classifier for as few times as possible.

\noindent\textbf{Stationary distribution.} To meet these requirements, the stationary distribution is designed as:
\begin{equation}
    \pi(x|\tilde{y})\propto LM(x)\cdot C(\tilde{y}|x) 
\end{equation}
where $LM(x)$ is the probability of the sentence ($x$) given by a pre-trained language model (LM) and $C(\tilde{y}|x)$ is the probability of an erroneous label ($\tilde{y}$) given by the victim model. $LM(x)$ guarantees fluency, while $C(\tilde{y}|x)$ is the attack target. 

\noindent\textbf{Transition proposal.} There are three word-level transition operations -- replacement, insertion and deletion. 
\textbf{Traversal indexing} is applied to select words on which operations are performed. Suppose \method selects the $i$-th word ($w_i$) on the $t$-th proposal, then on the $(t+1)$-th proposal, the selected word ($w^*$) is:

\begin{align}
w^*=\begin{cases}
        w_{i+1}, & if\ i\ne n \\
        w_1, & otherwise
    \end{cases} \notag
\end{align}

The transition function for \textbf{replacement} is as Equation \ref{equation:cgmh-bb replace transition}, where $w_m$ is the selected word to be replaced, and $\mathcal{Q}$ is a pre-selected candidate set, which will be explained later.
The \textbf{insertion} operation ($T_{i}^B(x'|x)$) consists of two steps -- inserting a random word into the position and then performing replacement upon it. The \textbf{deletion} operation is rather simple. $T_{d}^B(x'|x)=1$ if $x'=x_{-m}$, where $x_{-m}$ is the sentence after deleting the $m$-th word ($w_m$), or $T_{d}^B(x'|x)=0$ otherwise.

\begin{align}
  &T_{r}^B(x'|x)=\mathcal{I}\{w^c\in \mathcal{Q}\}\cdot \label{equation:cgmh-bb replace transition}  \\
  &\frac{\pi(w_1,\cdots,w_{m-1},w^c,w_{m+1},\cdots,w_n|\tilde{y})}{\sum_{w\in\mathcal{Q}}\pi(w_1,\cdots,w_{m-1},w,w_{m+1},\cdots,w_n|\tilde{y})} \notag
\end{align}

The proposal distribution is a weighted sum of the transition functions:

\begin{equation}
    g(x'|x)=p_rT_{r}^B(x'|x)+p_iT_{i}^B(x'|x)+p_dT_{d}^B(x'|x) \notag
\end{equation}

where $p_r$, $p_i$ and $p_d$ are pre-defined probabilities of the operations.

\paragraph{Pre-selection.} The \textbf{pre-selector} generates a candidate set for $T_r^B(x'|x)$ and $T_i^B(x'|x)$. It chooses the most possible words according to the score ($S^B(w|x)$) to form the candidate word set $\mathcal{Q}$. $S^B(w|x)$ is formulated as:

\begin{align}
    S^B(w|x)=LM(w|x_{[1:m-1]})\cdot LM_b(w|x_{[m+1:n]}) \notag
\end{align}

where $x_{[1:m-1]}=\{w_1,\cdots,w_{m-1}\}$ is the prefix of the sentence, $x_{[m+1:n]}$ is the suffix of the sentence, and $LM_b$ is a pre-trained backward language model. Without pre-selection, $\mathcal{Q}$ will include all words in the vocabulary, and the classifier will be invoked repeatedly to compute the denominator of Equation~\ref{equation:cgmh-bb replace transition}, which is inefficient.

\subsection{White-Box Attack}
\label{section:methods:subsection:mh-wb}

The only difference between white-box attack (\wbmethod) and \bbmethod lies in the pre-selector.


\noindent\textbf{{Pre-selection.}} In \wbmethod, the gradient is introduced into the pre-selection score ($S^W(w|x)$). $S^W(w|x)$ is formulated as:

\begin{align}
    S^W(w|x) = S^B(w|x)\cdot S(\frac{\partial\tilde{\mathcal{L}}}{\partial e_m},e_m-e) \notag
\end{align}

where $S$ is the cosine similarity function, $\tilde{\mathcal{L}}=\mathcal{L}(\tilde{y}|x,C)$ is the loss function on the target label, $e_m$ and $e$ are the embeddings of the current word ($w_m$) and the substitute ($w$). The gradient ($\frac{\partial\tilde{\mathcal{L}}}{\partial e_m}$) leads to the steepest direction, and $e_m-e$ is the actual changing direction if $e_m$ is replaced by $e$. The cosine similarity term ($S(\frac{\partial\tilde{\mathcal{L}}}{\partial w_m},\Delta w)$) guides the samples to jumping along the direction of the gradient, which raises $C(\tilde{y}|x)$ and $\alpha(x'|x)$, and eventually makes \wbmethod more efficient.

Note that insertion and deletion are excluded in \wbmethod, because it is difficult to compute their gradients. Take the insertion operation for instance. One may apply a similar technique in \bbmethod, by first inserting a random word forming intermediate sentence $x^*=\{w_1,\cdots,w_m,w^*,w_{m+1},\cdots,w_n\}$ and then performing replacement operation upon $x^*$. Computing $\frac{\partial\mathcal{L}(\tilde{y}|x^*,C)}{\partial w^*}$ is easy, but it is not the actual gradient. Computing of the actual gradient ($\frac{\partial\mathcal{L}(\tilde{y}|x,C)}{\partial w^*}$) is hard, since the change from $x$ to $x^*$ is discrete and non-differential.

%% file: experiments.tex
\section{Experiments}
\label{section:experiments}





\noindent\textbf{Datasets.} Following previous works, we validate the performance of proposed \method on IMDB and SNLI datesets.
The IMDB dataset includes 25,000 training samples and 25,000 test samples of movie reviews, tagged with sentimental labels (positive or negative). 
The SNLI dataset contains 55,000 training samples, 10,000 validation samples and 10,000 test samples.
Each sample contains a premise, a hypothesis and an inference label (entailment, contradiction or neutral). 
We adopt a single layer bi-LSTM and the BiDAF model~\citep{seo2016bidirectional} (which employs bidirectional attention flow mechanism to capture relationships between sentence pairs) as the victim models on IMDB and SNLI, respectively.


\noindent\textbf{Baseline Genetic Attacker.} We take the state-of-the-art genetic attack model~\citep{alzantot2018generating} as our baseline, which uses a gradient-free population-based algorithm. 
Intuitively, it maintains a population of sentences, and perturbs them by word-level replacement according to the embedding distances without considering the victim model. 
Then, the intermediate sentences are filtered by the victim classifier and a language model, which leads to the next generation.

\noindent\textbf{Hyper-parameters.} As in the work of \newcite{miao2018cgmh}, \method is limited to make proposals for at most 200 times, and we pre-select 30 candidates at each iteration. 
Constraints are included in \method to forbid any operations on sentimental words (eg. ``great'') or negation words (eg. ``not'') in IMDB experiments with SentiWordNet~\citep{esuli2006sentiwordnet,baccianella2010sentiwordnet}.
All LSTMs in the victim models have 128 units. The victim model reaches 83.1\% and 81.1\% test accuracies on IMDB and SNLI, which are acceptable results. More detailed hyper-parameter settings are included in the appendix.

\begin{figure}[t]
    \centering
    \subfigure[IMDB]{
        \includegraphics[width=0.46\columnwidth]{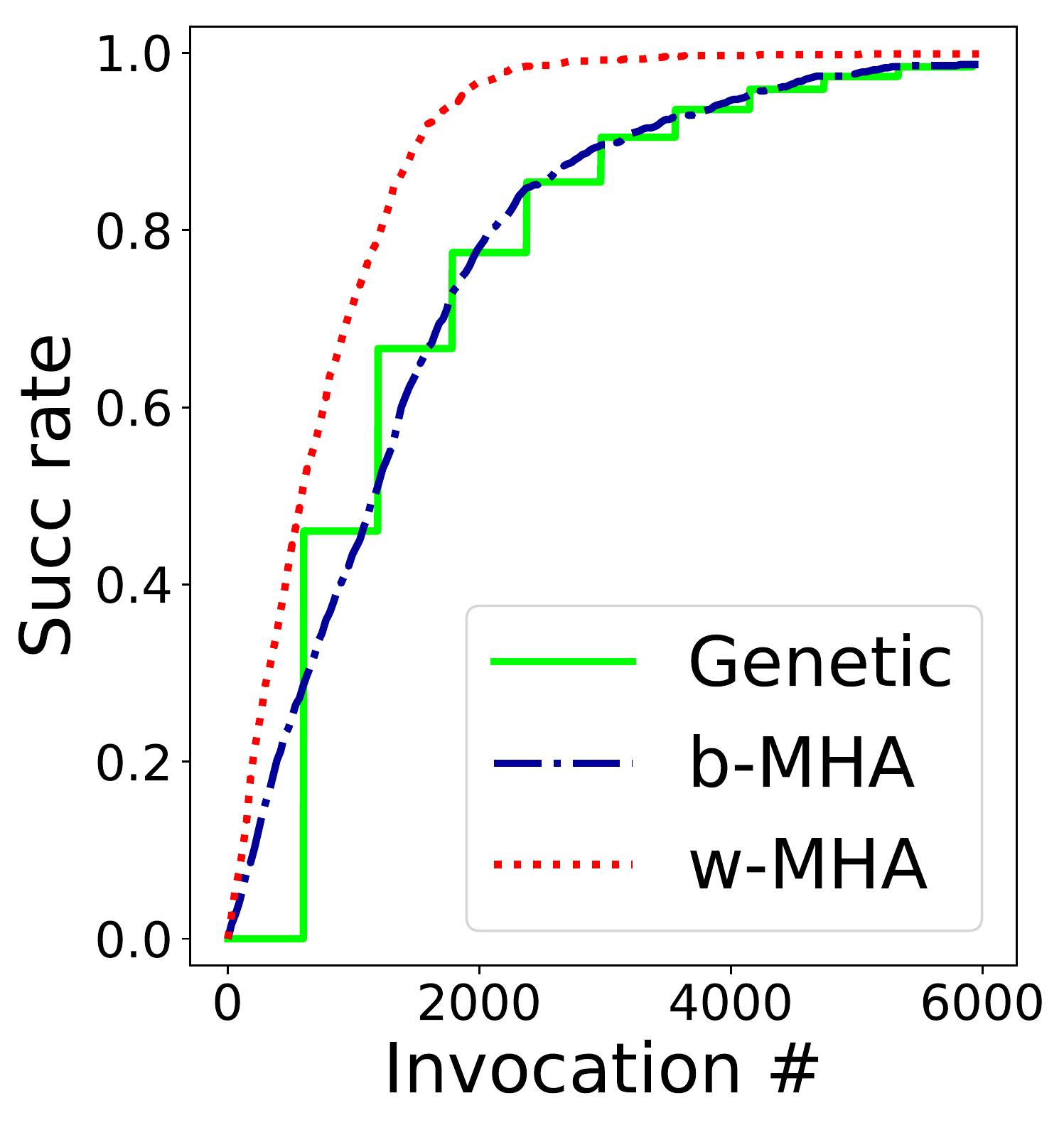}
    }
    \subfigure[SNLI]{
        \includegraphics[width=0.46\columnwidth]{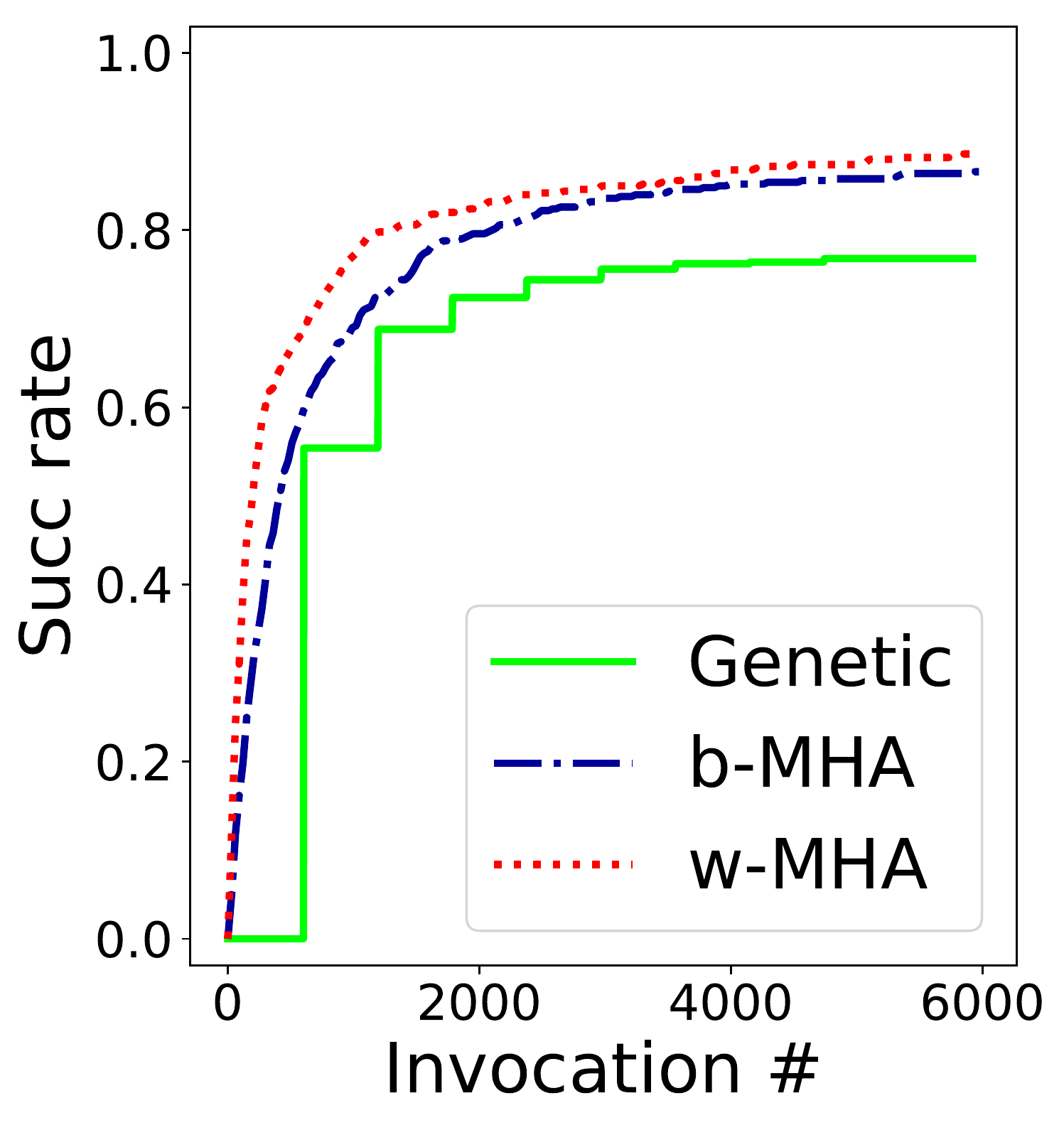}
    }
    \caption{Invocation-success curves of the attacks.}
    \label{figure:invok succ}
\end{figure}

\begin{table}[t]\footnotesize
    \centering
    \begin{threeparttable}
    \begin{tabular}{cccccc}
        \toprule
Task & Approach & Succ(\%) & Invok\# & PPL & $\alpha$(\%)\tnote{} \\
        \midrule
\multirow{3}{*}{\rotatebox{90}{IMDB}} & Genetic & 98.7 & 1427.5 & 421.1 & -- \\
& \bbmethod & 98.7 & 1372.1 & 385.6 & 17.9 \\
& \wbmethod & \textbf{99.9} & \textbf{748.2} & \textbf{375.3} & 34.4 \\
\cline{1-6}
\multirow{3}{*}{\rotatebox{90}{SNLI}} & Genetic & 76.8 & 971.9 & 834.1 & -- \\
& \bbmethod & 86.6 & 681.7 & 358.8 & 9.7 \\
& \wbmethod & \textbf{88.6} & \textbf{525.0} & \textbf{332.4} &13.3 \\
        \bottomrule
    \end{tabular}
    \end{threeparttable}
    \caption{Adversarial attack results on IMDB and SNLI. The acceptance rates~($\alpha$) of M-H sampling are in a reasonable range.}
    \label{table:attack}
\end{table}

\begin{table}[t]\footnotesize
    \centering
    \begin{threeparttable}
    \begin{tabular}{p{0.45\textwidth}}
        \toprule Case 1 \\\cline{1-1}
\textbf{Premise:} \emph{three men are sitting on a beach dressed in orange with refuse carts in front of them.} \\
\textbf{Hypothesis:} \emph{empty trash cans are sitting on a beach.} \\
\textbf{Prediction:} $\langle$Contradiction$\rangle$ \\
\cline{1-1}
\textbf{Genetic:} \emph{\textbf{\color{red}empties} trash cans are sitting on a beach.} \\
\textbf{Prediction:} $\langle$Entailment$\rangle$ \\
\cline{1-1}
\textbf{\bbmethod:} \emph{\textbf{\color{brown}the} trash cans are sitting \textbf{\color{brown}in} a beach.} \\
\textbf{Prediction:} $\langle$Entailment$\rangle$ \\
\cline{1-1}
\textbf{\wbmethod:} \emph{\textbf{\color{blue}the} trash cans are sitting on a beach.} \\
\textbf{Prediction:} $\langle$Entailment$\rangle$ \\
        \midrule Case 2\\\cline{1-1}
\textbf{Premise:} \emph{a man is holding a microphone in front of his mouth.} \\
\textbf{Hypothesis:} \emph{a male has a device near his mouth. } \\
\textbf{Prediction:} $\langle$Entailment$\rangle$ \\
\cline{1-1}
\textbf{Genetic:} \emph{a \textbf{\color{red}masculine} has a device near his mouth.} \\
\textbf{Prediction:} $\langle$Neutral$\rangle$ \\
\cline{1-1}
\textbf{\bbmethod:} \emph{a man has a device near his \textbf{\color{brown}car}.} \\
\textbf{Prediction:} $\langle$Neutral$\rangle$ \\
\cline{1-1}
\textbf{\wbmethod:} \emph{a man has a device near his \textbf{\color{blue}home}.} \\
\textbf{Prediction:} $\langle$Neutral$\rangle$ \\
        \bottomrule
    \end{tabular}
    \end{threeparttable}
    \caption{Adversarial examples generated on SNLI.}
    \label{table:cases}
\end{table}

\subsection{Adversarial Attack}
\label{section:experiments:subsection:adv attack}

To validate the attacking efficiency, we randomly sample 1000 and 500 correctly classified examples from the IMDB and SNLI test sets, respectively.
Attacking success rate and invocation times~(of the victim model) are employed for testing efficiency.
As shown in Figure \ref{figure:invok succ}, curves of our proposed \method are above the genetic baseline, which indicates the efficiency of \method.
By incorporating gradient information in proposal distribution, \wbmethod even performs better than \bbmethod, as the curves rise fast. 
Note that the ladder-shaped curves of the genetic approach is caused by its population-based nature.

We list detailed results in Table \ref{table:attack}.
Success rates are obtained by invoking the victim model for at most 6,000 times.
As shown, the gaps of success rates between the models are not very large, because all models can give pretty high success rate.
However, as expected, our proposed \method provides lower perplexity~(PPL)~\footnote{We use the open released GPT2~\citep{radfordlanguage} model for PPL evaluation.}, which means the examples generated by \method are more likely to appear in the corpus of the evaluation language model. As the corpus is large enough and the language model for evaluation is strong enough, it indicates the examples generated by \method are more likely to appear in natural language space. It eventually leads to better fluency.

Human evaluations are also performed. From the examples that all three approaches successfully attacked, we sample 40 examples on IMDB. Three volunteers are asked to label the generated examples. Examples with false labels from the victim classifier and with true labels from the volunteers are regarded as actual adversarial examples. The adversarial example ratios of the genetic approach, \bbmethod and \wbmethod are 98.3\%, 99.2\% and 96.7\%, respectively, indicating that almost all generated examples are adversarial examples. Volunteers are also asked to rank the generated examples by fluency on SNLI (``1'' indicating the most fluent while ``3'' indicating the least fluent). 20 examples are sampled in the same manners mentioned above. The mean values of ranking of the genetic approach, \bbmethod and \wbmethod are 1.93, 1.80 and 2.03, indicating that \bbmethod generates the most fluent samples.
Samples generated by \wbmethod are less fluent than the genetic approach. It is possibly because the gradient introduced into the pre-selector could influence the fluency of the sentence, from the perspective of human beings.

Adversarial examples from different models on SNLI are shown in Table \ref{table:cases}. The genetic approach may replace verbs with different tense or may replace nouns with different plurality, which can cause grammatical mistakes (\emph{eg.} Case 1), while \method employs the language model to formulate the stationary distribution in order to avoid such grammatical mistakes. \method does not have constraints that word replacement should have similar meanings. \method may replace entities or verbs with some irrelevant words, leading to meaning changes of the original sentence (\emph{eg.} Case 2). More cases are included in the appendix.

\begin{table}[t]\footnotesize
    \centering
    \begin{threeparttable}
    \begin{tabular}{lccc}
        \toprule
\multirow{2}{*}{Model} & \multicolumn{3}{c}{Attack succ (\%)} \\
\cline{2-4}
 & Genetic & \bbmethod & \wbmethod \\
        \midrule
Victim model & 98.7 & 98.7 & 99.9 \\
+ Genetic adv training & 93.8 & 99.6 & 100.0 \\
+ \bbmethod adv training & 93.0 & 95.7 & 99.7 \\
+ \wbmethod adv training & 92.4 & 97.5 & 100.0 \\
        \bottomrule
    \end{tabular}
    \end{threeparttable}
    \caption{Robustness test results on IMDB.}
    \label{table:robustness}
\end{table}

\begin{table}[t!]\small
    \centering
    \begin{threeparttable}
    \begin{tabular}{lccc}
        \toprule
\multirow{2}{*}{Model} & \multicolumn{3}{c}{Acc (\%)} \\
\cline{2-4}
& Train \# = 10K & 30K & 100K \\
        \midrule
Victim model & 58.9 & 65.8 & 73.0 \\
+ Genetic adv training & 58.8 & 66.1 & \textbf{73.6} \\
+ \wbmethod adv training & \textbf{60.0} & \textbf{66.9} & \textbf{73.5} \\
        \bottomrule
    \end{tabular}
    \end{threeparttable}
    \caption{Accuracy results after adversarial training.}
    \label{table:adv train}
\end{table}

\subsection{Adversarial Training}
\label{section:experiments:subsection:adv training}

In order to validate whether adversarial training is helpful for improving the adversarial robustness or classification accuracy of the victim model, a new model is trained from scratch after mixing the generated examples into the training set. 

To test the adversarial robustness, we attack the new models with all methods on IMDB. As shown in Table \ref{table:robustness}, the new model after genetic adversarial training can not defend \method. On the contrary, adversarial training with \bbmethod or \wbmethod decreases the success rate of genetic attack.
It shows that the adversarial examples from \method could be more effective than unfluent ones from genetic attack, as assumed in Figure \ref{figure:manifold}.

To test whether the new models could achieve accuracy gains after adversarial training,
experiments are carried out on different sizes of training data, which are subsets of SNLI's training set. The number of adversarial examples is fixed to 250 during experiment.
The classification accuracies of the new models after the adversarial training by different approaches are listed in Table \ref{table:adv train}.
Adversarial training with \wbmethod significantly improves the accuracy on all three settings~(with p-values less than 0.02).
\wbmethod outperforms the genetic baseline with 10K and 30K training data, and gets comparable improvements with 100K training data.
Less training data leads to larger accuracy gains, and \method performs significantly better than the genetic approach on smaller training set.


%% file: summary.tex
\section{Future Works}
\label{section:future works}

Current \method returns the examples when the label is changed, which may lead to incomplete sentences, which are unfluent from the perspective of human beings. Constraints such as forcing the model to generate $\langle$EOS$\rangle$ at the end of the sentence before returning may address this issue.

Also, entity and verb replacements without limitations have negative influence on adversarial example generations for tasks such as NLI. Limitations of similarity during word operations are essential to settle the problem. Constraints such as limitation of the embedding distance may help out. Another solution is introducing the inverse of embedding distance in the pre-selection source.

\section{Conclusion}
\label{section:summary}

In this paper, we propose \method, which generates adversarial examples for natural language by adopting the MH sampling approach.
Experimental results show that our proposed \method could generate adversarial examples faster than the genetic baseline.
Obtained adversarial examples from \method are more fluent and may be more effective for adversarial training.


\section{Acknowledgments}

We would like to thank Lili Mou for his constructive suggestions. We also would like to thank the anonymous reviewers for their
insightful comments.